\def\year{2020}\relax
\documentclass[letterpaper]{article} 
\usepackage{aaai20}  
\usepackage{times}  
\usepackage{helvet} 
\usepackage{courier}  
\usepackage[hyphens]{url}  
\usepackage{graphicx} 
\usepackage{tabularx}
\usepackage{amssymb}
\usepackage[cmex10]{amsmath}
\usepackage{mathrsfs}
\usepackage{euscript}
\usepackage{float}
\usepackage[numbers,sort]{natbib}
\let\OLDthebibliography\thebibliography
\renewcommand\thebibliography[1]{
  \OLDthebibliography{#1}
  \setlength{\parskip}{0pt}
  \setlength{\itemsep}{0pt plus 0.1pt}
}
\usepackage{diagbox}
\usepackage{caption,subfig}
\usepackage[figuresright]{rotating}
\usepackage{multirow}

\usepackage{array}
\newcolumntype{L}{>{\centering\arraybackslash}m{2.7cm}}
\graphicspath{{images/}{use/}{infer/}}
\usepackage[algo2e,linesnumbered,ruled,vlined]{algorithm2e} 
\newcommand{\nosemic}{\renewcommand{\@endalgocfline}{\relax}}
\newcommand{\dosemic}{\renewcommand{\@endalgocfline}{\algocf@endline}}
\usepackage{csquotes}
\makeatletter
\newcommand{\removelatexerror}{\let\@latex@error\@gobble}
\makeatother
\newcommand{\quotes}[1]{``#1''}

\usepackage{url}
\usepackage{enumerate}

\title{Online Similarity Learning with  Feedback for Invoice Line Item Matching }
\author{ \Large \textbf{ \textsuperscript{\rm 1} Chandresh Kumar Maurya, } \textbf{\textsuperscript{\rm 2} Neelamadhav Gantayat, \textsuperscript{\rm 2} Sampath Dechu, \textsuperscript{\rm 1} Tom\'a\v{s} Horv\'ath}\\ 
\textsuperscript{\rm 1} Department of Data Science and Engineering, T-Labs, E\"{o}tv\"{o}s Lor\'and University, Budapest, Hungary \\ 
\textsuperscript{\rm 2} IBM Research AI, Bangalore, India \\
\{ckm.jnu, neelamadhavg, sampath.dechu\}@gmail.com, tomas.horvath@inf.elte.hu 
}
 
\begin{document}

\maketitle

\begin{abstract}
The procure to pay process (P2P) in large enterprises is a back-end business process which deals with the procurement of products and services for enterprise operations. Procurement is done by issuing purchase orders to impaneled vendors and invoices submitted by vendors are paid after they go through a rigorous validation process. Agents orchestrating P2P process often encounter the problem of matching a product or service descriptions in the invoice to those in purchase order and verify if the ordered items are what have been supplied or serviced.  For example, the description in the invoice and purchase order could be  {\it TRES 739mL CD KER Smooth} and {\it TRES 0.739L CD KER Smth} which look different at word level but refer to the same item. In a typical P2P process, agents are asked to manually select the products which are similar before invoices are posted for payment. This step in the business process is manual, repetitive, cumbersome, and costly. Since descriptions are not well-formed sentences, we cannot apply existing semantic and syntactic text similarity approaches directly. In this paper, we present two approaches to solve the above problem using various types of available agent's recorded feedback data.  If the agent's feedback is in the form of a relative ranking between descriptions, we use similarity ranking algorithm. If the agent's feedback is absolute such as match/no-match, we use classification similarity algorithm. We also present the threats to the validity of our approach and present a possible remedy making use of product taxonomy and catalog. We showcase the comparative effectiveness and efficiency of the proposed approaches over many benchmarks and real-world data sets. 
\end{abstract}
\section{Introduction}
\begin{figure}
\centering
\includegraphics[height=4cm]{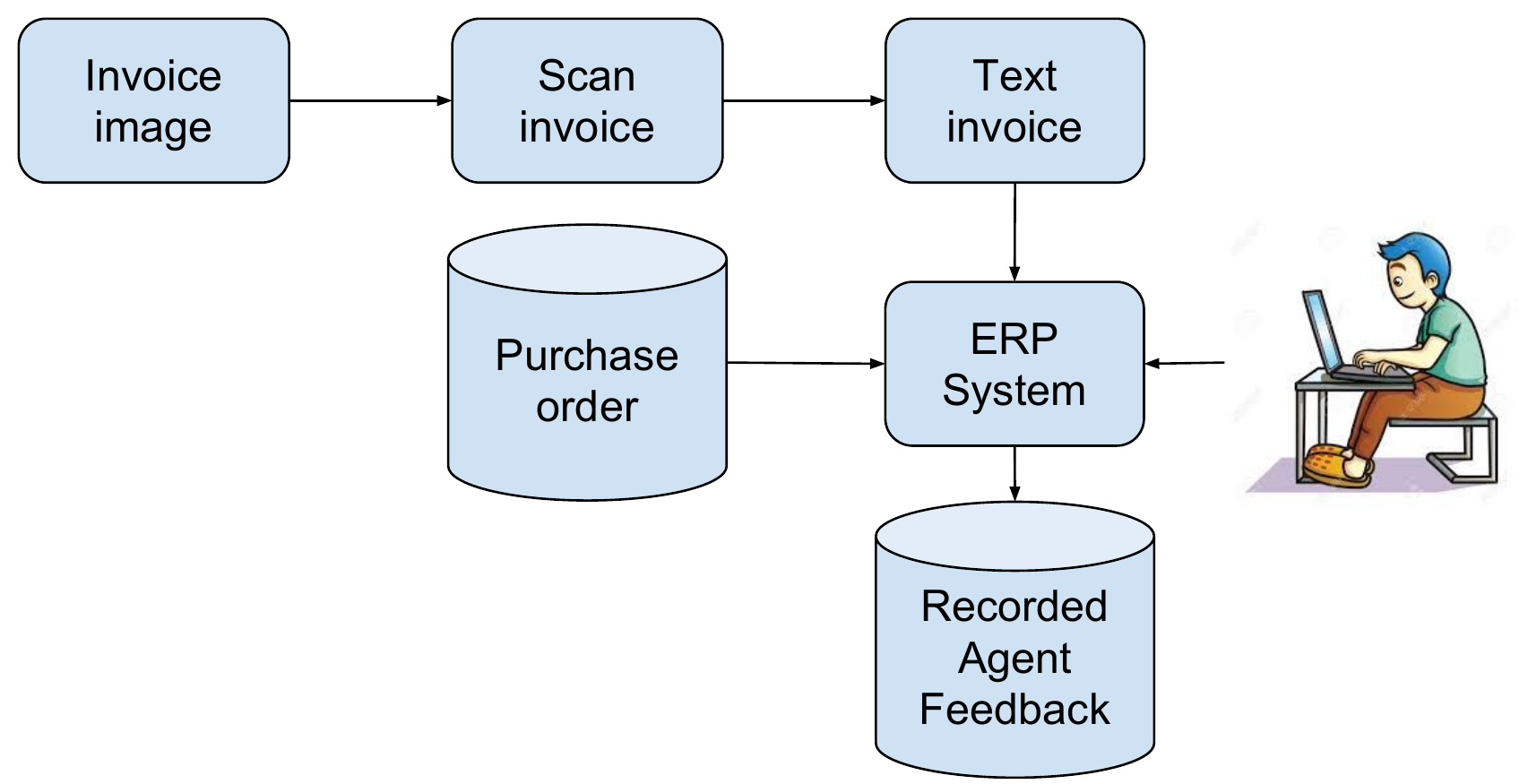}
\caption{Invoice line item matching in P2P business process}
\label{fig:invoice}
\end{figure}

\noindent 
Procure to pay process (P2P) is a resource-intensive business process in large enterprises. The process deals with procuring raw materials, services that are needed for enterprise operations from impaneled vendors and paying invoices submitted by the vendors within contractual payment deadlines. Invoices go through various processing steps such as validation of invoice, validation of vendor credentials, validation of tax compliance of transaction, etc. before they are posted for payment. One of the key steps before payments are made to the vendor is the invoice line item matching step. As depicted in Figure~\ref{fig:invoice}, during invoice line item matching, a domain expert compares the descriptions of the products or services to that of what was ordered in purchase order and validates if they refer to the same or similar products/services. These selections of matched descriptions between invoice and purchase order (PO) are recorded in ERP systems.  An example of descriptions in invoice and PO are given in Table \ref{lineitem}. The line items in the invoice column are similar to both the descriptions in the PO column, but only the first description in the PO column is the correct match to the invoice description. Invoice line item matching process is very cumbersome, repetitive and resource intensive. Currently, IBM processes more than 50 million invoices per year from third parties and employ thousands of agents to do the task. The time taken to process per invoice is usually 4-5 minutes by each agent and they have to process more than hundreds of invoices every day. There is a dire need to apply efficient machine learning techniques to reduce the cost and time to process the invoices and help the agent to become more efficient. 

 \begin{table}[]
\caption{An example of line item matching}
\label{lineitem}
\resizebox{\columnwidth}{!}{%
\begin{tabular}{|l|l|}
\hline
Invoice                           & PO                                                                                                                                        \\ \hline \hline

TRES 739mL CD KER Smooth  & \begin{tabular}[c]{@{}l@{}}1. TRES 0.739L CD KER Smth\\ 2. Tres Soya Smooth Conditioner 150 gm\end{tabular} \\ \hline

5x200ml Fruit Juice 100\% - Tropicana, Apple  & \begin{tabular}[c]{@{}l@{}}1. Tropicana 100\% Apple Juice -  1L\\ 2. Fruit Juice 500ml - Tropicana, Custard Apple\end{tabular} \\ \hline
Battery Distilled Water Replacement      & \begin{tabular}[c]{@{}l@{}}1. Battery Maintenance Services\\ 2.  Battery Warranty extension\end{tabular}                              \\ \hline
\end{tabular}}
\end{table}
 An invoice comes into the processing pipeline as an image. Image is scanned by an optical character recognition (OCR) software and textual representation of invoices is generated. Textual invoice is posted into the enterprise ERP software for matching against the purchase order (PO) with the agent intervention to allow or disallow the invoice to go further into the processing pipeline. The mismatches between invoice and PO descriptions occur due to (1) different ways of describing products by vendors and buyers, (2) slightly different products supplied by vendor, and (3) error caused by the OCR while scanning the invoice document.  The task of line item description matching appears close to the sentence similarity task, but there are significant differences due to domain-specific descriptions. We refer to this task as semantic textual description similarity (STDS) task. 
 
String matching and well-formed sentence matching approaches cannot be directly applied to STDS task due to lack of awareness to domain-specific knowledge agents apply while matching the descriptions. For example, 10 boxes of pencils vs 10 pencil boxes appear same at the syntactic level, but one refers to 10 boxes which contain pencils while another one refers to 10 boxes which are used to put pencils. Another challenge with the descriptions is the presence of out-of-vocabulary (OOV) words, spelling mistakes (OCR errors), acronyms, etc. in the description. Matching numeric quantities such as $739mL$ and $0.739L$ present in the string is also important. Writing rules to handle all such cases is not feasible due to the diversity of domains where invoice matching process is executed. Using big keyword table from buyers and suppliers is also not a viable option due to (1) the number of different items (e.g. Amazon has millions of product), and (2) OOV word, OCR errors, and acronyms make the table based matching error-prone.   Recent advances in deep learning have also been applied to semantic text similarity task for well-formed sentences by embedding the joint similarity into a latent space using character-level bidirectional LSTM \cite{neculoiu2016learning}. 

Another challenge in applying existing string matching approaches is considering the agent's online feedback.  Sometimes, the agent mentions their preferences about the descriptions that match and would approve the invoice for further steps in the P2P process. For example, the agent may rank (glycerine, glycerine 2\%) pair to be a better match than the pair (glycerine,  glycerine 20\%). It is not clear how to take into account such implicit feedback in string matching approaches at runtime. 
In another scenario, the feedback that is received from the agent can be binary in nature. For example, she might label the pair (glycerine, glycerine 2\%) as similar whereas  (glycerine, glycerine 20\%) as dissimilar. In the present work, we want to exploit such feedback to learn a similarity function. 

The main contribution of our work are as follows:
\begin{itemize}
 \itemsep0em
\item  Two approaches  are proposed to match descriptions using domain knowledge captured in  the user's feedback. First approach learns similarity rank when recorded users feedback has relative ranking of description matches and second approach uses binary classification when users' recorded feedback is absolute match/no-match between pair of descriptions. 
\item To circumvent the issue of hierarchical relationship among items in the invoice and PO line items, we present an algorithm that makes use of product taxonomy and catalog.

\item Proposed approaches are evaluated on real-world description datasets e.g. invoice data from internal clients, publicly available product description datasets  and compare the results with the state-of-the-art approaches applied to natural language sentences and show limitations of the existing work. Additionally, we also evaluate the proposed approachs using different kinds of {\it description representations} such as character n-gram (n=2 to 5), directly encoding the description assuming them as sentence via pre-trained models from Infersent \cite{conneau-EtAl:2017:EMNLP2017} and Google's universal sentence encoder.  \cite{cer2018universal}.
\end{itemize}


\section{Related Work}\label{related}
Similarity learning has been studied extensively in data mining and machine learning communities \cite{chechik2010large,crammer2012adaptive,neculoiu2016learning,hao2015learning,pawar2018calculating,li2006sentence}.
Because of the vast literature available, here we present the most relevant works. In \cite{chechik2010large}, the author presents an online relative similarity learning task for images. They learn a metric $W$ based on the triples of the images within the passive-aggressive learning framework \cite{crammer2006online}. Along the same line of work \cite{gao2017sparse} presents an extension to handle the sparsity present in the image data for image retrieval task. These techniques are developed for image similarity task. It is not clear how well these methods will perform if employed in the text domain. In \cite{liu2015computing}, the author uses support vector regression with various features such as WordNet-Based features, corpus-based features, Word2Vec-based feature, Alignment-based features, and Literal-based features to predict the similarity between short English sentences. Though the method is able to capture various aspects of the sentences, the proposed method needs large corpora and runs offline. 
Deep learning based approaches such as \cite{neculoiu2016learning} uses the character level bidirectional LSTM to learn the semantic similarity of the strings. They claim to handle abbreviations, misspellings, extra words, annotations, typos, etc. The three problems in using such methods are: (1) they need large parallel corpora,(2) huge training time, and (3) not catering user's feedback while training. 

Recent works such as \cite{pawar2018calculating,li2006sentence} try to capture semantic similarity via linguistic resources. However, the accuracy of these models is very sensitive to spelling mistakes. For example, the similarity score for the sentence pair (soft butter, soft butter) from the first model is  0.970. On the other hand, if there is just a single spelling mistake such as \emph{butter} becomes \emph{buttre}, similarity score drops to 0.0835. To account for such spelling mistakes, we include lexical normalization of the noun phrases so that the final performance of the system is not hampered too much.  In \cite{huai2018uncorrelated}, a method is proposed to learn a metric which can be used to learn the patient similarity based on a complex medical record. There also exists work in e-commerce and recommendation system domain for similarity learning. For example, in \cite{kutiyanawala2018towards,hu2018reinforcement} the author propose matching query to items in the product catalog.  Most of these techniques rely on the availability of the product catalog.
\section{Proposed Approach}\label{system}
In the setting we are solving the problem, invoices come in a streaming fashion that means an invoice comes to the agent for matching line item against the line item in the PO and give feedback in terms of the relative ranking or absolute ranking. We do some pre-processing beforehand and then present it to the agent for her feedback.  
The proposed algorithm as presented in Algorithm~\ref{algo1} is called Similarity Learning with Feedback (SLF). While using Ranking similarity, we abbreviate it as SLFR and when we refer to classification similarity, we abbreviate it as SLFC. Due to lack of space, we provide a brief description of each step in Algorithm \ref{algo1}. We also skip further details of the SLFC algorithm for the same reason. 

\begin{algorithm2e}[tb]
	\SetAlgoLined\DontPrintSemicolon
	{\small
		\KwIn{Aggressiveness parameter C/learning rate $\eta$,
		Invoice and PO line items.
		}{
		\KwOut{$ W_{T}$}.
		 \label{algo1}
      {\bf Initialize:} $W_0 = I (\text{identity Matrix})$ or weight vector ${\bf w}_0 = \boldsymbol{0}$.\\
    \For{$t := 1,...,T$}
    {
      Apply Lexical Normalization as discussed in to the query string (e.g. Invoice string).\\
      Receive $K$ strings via fuzzy matching from the pool for query string $s$.\\
    Extract noun phrases from string pair. If noun phrases did not match, return fuzzy matching score 0.\\
    
    Present the pair of strings $(s, s_i)$ to the agent where $s_i$ is the best fuzzy matching string.\\
    if the agent did not like the pair and gives negative vote, randomly sample a string $s_j$ from the remaining pool of strings.\\
    if the agent prefers the pair $(s, s_j)$ more than the pair $(s, s_i)$, we form triple of strings $(s, s_i, s_j)$. If the agent labels pair $(s, s_i)$ as dissimilar and the pair $(s, s_j)$ as similar, we form data for binary classification.\\
    {\bf Update:}
          \begin{equation*}
               \text{Ranking Similarty}
               \begin{cases}
               W_{t+1} = W_t + \tau_tU_t \\
               \tau_t = \min({C, \ell^1_t/\|U_t\|^2})\\
               U_t=[s^1(s_j-s_i)..s^d(s_j-s_i)]^T
               \end{cases}
          \end{equation*}
          \hspace{4cm} OR
        
         \text{Classification Similarity}
            
    }
    }
	}
	\caption{Similarity Learning with Feedback (SLF) Algorithm}\label{algo1}
\end{algorithm2e}



In short, Algorithm \ref{algo1} consists of the following steps:
\begin{itemize}
    \item {\bf Lexical Normalization }In the first step, we apply lexical normalization as given in \cite{han2011lexical} to the query string (e.g. invoice string). Lexical normalization is the process of detecting \quotes{ill-formed OOV} words. For example, people often misspell \quotes{glycerine} to \quotes{glycerien}. Otherwise existing knowledge-based approaches will not be able to detect proper nouns. For every query string $s$, we return top $K$ fuzzy matching strings from the pool.  The fuzzy matching is calculated by cosine similarity between bi-grams and tri-grams representations. 
    \item {\bf Noun Phrase Extraction} 
    For avoiding matching invoice items such as 
    \quotes{Glycerine white distilled 12\%} and \quotes{Vinegar white distilled 12\%}  we need to remove such string pairs from further comparison. We use python spacy \footnote{https://spacy.io/} API for noun phrase chunking.
    \item {\bf Agent preference}
    The best fuzzy matching string $s_i$ is paired with the query string $s$ and presented to the agent. The reason behind presenting the best fuzzy matching strings is to assist the agent and save time in processing invoices.     If the agent allows the pair to go through the P2P system, this means that fuzzy matching is performing well. On the other hand, if the agent dislikes the pair, she is presented with another string $s_j$ selected uniformly and randomly  or second best from the remaining pool of strings. The idea behind choosing next best fuzzy matching string logically make sense since among all the matching strings, next best fuzzy matching string is suppose to be the second best match for the invoice string. Random choice of strings may be used in scenario when all strings except the best one score poorly on fuzzy score. If the agent prefers the pair $(s, s_j)$ more than the pair $(s, s_i)$, we collect such triples and train a ranking similarity model as discussed below. On the other hand, if the agent labels pair $(s,s_i)$ as dissimilar and the pair $(s, s_j)$ as similar, we train a binary classification model discussed below. Further, note that agent feedback is used here to collect as many triples as possible to form training data. As we show in experiment section that precision improves overtime as we receive more training data. 
    \item {\bf Ranking similarity model}
    Fuzzy matching can not be relied completely as it may not always return satisfactory result since it does not take into account the agent's preference. Hence, the idea is to learn from triples of string where it does not perform well and learn a ranking similarity model.  We present such a model that captures the agent's preference over  time and eventually may replace the fuzzy matching altogether.
    
    The model that we use for ranking similarity comes from metric learning literature \cite{chechik2010large}. 
    We want to learn a function $f(\cdot, \cdot)$ that assigns high score to pairs $(s, s_j)$ than the pair $(s, s_i)$ whenever the agent prefers $(s, s_j)$ more than $(s, s_i)$. Assume that the function $f$ has a bilinear form shown in \eqref{bilinear}.
    \begin{equation}\label{bilinear}
        f_W(s_i, s_j) := s_i^TWs_j
    \end{equation}
    where the matrix $W \in R^{d\times d}$ and $d$ is the dimensionality of the feature vector. Note that the strings and its representation both are denoted by the column vector $s \in R^d$. Further this model is put under PA learning framework \cite{crammer2006online} and solved via forming Lagrangian. Further details of the algorithm is the same as in \cite{chechik2010large}.  The update equations are shown in step 7 of the algorithm 1.  Updating $W$ efficiently in memory can be challenging since it is a huge dense matrix when $d$ is large. In \cite{hao2015learning}, the author proposes to use $L_1$ regularization to generate sparse solution and {\it diagnolisation trick} to save memory foot-print. However, empirical results on the performance metrics and time taken to compute optimal $W$ indicate that there is a huge trade-off between the performance gain achieved using $L_1$ regularization and time-complexity (some empirical results finished in 9 days compared to the OASIS method which finished in 3 days). Therefore, we do not use the $L_1$ regularization for sparse solution though this can indeed be plugged in here.
\end{itemize}
 \section{Datasets and Preprocessing}\label{dataset}
To demonstrate the effectiveness and efficiency of the proposed approaches, the following datasets are used for empirical comparison as shown in  Table \ref{data}.
\begin{table}[htbp]
\centering
\caption{Summary of  datasets  used in the experiment}
\label{data}
\resizebox{\columnwidth}{!}{%
\begin{tabular}{|l|l|l|l|}
\hline
Dataset & \#Train &\#Test  & \#Features   \\ \hline \hline
Invoice & 370 & 184 & 3649\\\hline
Amazon Electronics & 9368 &4683 & 35327\\\hline
Amazon Automative& 21107 & 10553 & 40123\\\hline
Amazon Home&21887 & 10943 &46453\\\hline
Flipkart & 9417 & 4708 & 19400\\\hline
SNLI & 121895& 60947&55956\\\hline
SICK & 3865 & 1932 & 18379\\\hline
STS & 1426 & 713 & 16110\\\hline
\end{tabular}}
\end{table}
We briefly describe the datasets and how do we create second/third string  in the absence of them for some datasets. Invoice data is a small  and real world data coming from a client. STS \footnote{http://alt.qcri.org/semeval2014/task3/index.php?id=data-and-tools}, SICK\footnote{http://alt.qcri.org/semeval2014/task1/}, and SNLI\footnote{https://nlp.stanford.edu/projects/snli/} are benchmark datasets used in similarity learning task. Amazon \footnote{https://github.com/SamTube405/Amazon-E-commerce-Data-set/tree/master/Data-sets} and Flipkart \footnote{https://data.world/promptcloud/product-details-on-flipkart-com} are product catalog data from  Amazon and Flipkart e-commerce sites. 
\subsection{Data Preprocessing} In this section, we discuss how to prepare the data for ranking similarity task. For ranking similarity task, we need triples of strings. We discuss how to derive such triples for each of the datasets used. 

Invoice data consists of invoice strings ($s$). We have the corresponding PO strings ($s_j$) a.k.a second string) as well. Since there is no third string ($s_i$) available, we manually curated and generated the third string from the second one (PO string) under the following assumption so that third string is less similar to the invoice string compared to the PO string. The following rules are derived during manual curation of the invoice data:
\begin{enumerate}
    \item Common antonyms such as men vs women.
    \item Small $\delta$ Numeric addition or deletion - for second string
    \item Large $\delta$ Numeric addition or deletion - for third string
    \item Replace Brand names, if applicable
    \item Replace Product names, if applicable
    \item String Manipulation such as insertion/deletion/substitution of random character and shuffle words
\end{enumerate}
Above rules are derived by observing some POs containing multiple strings (item description). 
In practice, there are multiple PO strings that can match with the invoice string, we assume that the PO string available to us has been returned by the Fuzzy matching API in the absence of the multiple PO string. In order to make sure that the above rules hold true (fuzzy matching API scores high for PO string), we also manually curated PO string to look more similar to invoice string than the third string derived from PO string. An example of string triple from invoice data look like as follows:
\begin{itemize}
    \item [$s$:]12z Dove Men US 2in1 FRts
   \item [$s_j$:] 11z Dove Men US 2in1 Frts 
 \item [$s_i$:] 12z Dove women US 2in1 Shampoo
\end{itemize}

STS, SICK, and SNLI data are benchmark data for semantic similarity task. These data consist of a string pair labeled as similar or not. We pick one of the strings from these string pair and apply insertion/deletion/substitution as well as random string concatenation transformation to form the third string. Amazon and Flipkart data has just the product description. To generate the second string, we compulsorily apply rule 4 and randomly apply rule 2 and 6. Doing so preserves the main product. For example, \quotes{sanoxy analog to digital audio converter adapter} changes to \quotes{Philips analog to digital audio converter adapter}. For the third string, we compulsorily apply rule 3 and randomly apply rule 1, 5, and 6. This transformation preserves the brand name but changes the product. For example, \quotes{sanoxy analog to digital audio converter adapter} changes to \quotes{sanoxy analog to digital audio converter TV}. 
\subsection{Feature Representation}
Ranking  similarity tasks require numeric feature vector representation. Therefore, we tried out various encoding schemes for the invoice and PO strings such as tf-idf, sentence embedding using Facebook Infersent and Google's universal sentence encoder. For each string, we consider character n-gram (n=2 to 5) as well as word level n-gram to compute the tf-idf vector. Feature vector size using tf-idf encoding scheme is shown in the Table \ref{data} (\#Features column).

We also want to see how good is sentence embedding compared to vanilla tf-idf based feature representation. For sentence embedding, we use pre-trained model from Infersent which is trained on Glove word-vectors and Google's universal sentence embedding(USE) which is trained on variety of datasets.  In our experiments, we observed that around 80\% of the words in the experimental dataset are present in the Infersent/USE embedding. The feature vector size using Infersent encoding is 4096 and using USE is 512 for all the datasets.

\section{Experimental Testbed and Setup}\label{exp}
The proposed algorithm is compared with some state-of-the-art algorithms in the literature. For relative ranking similarity task (SLFR), we compare against the following methods: 
\begin{itemize}
\item \textbf{Cosine Similarity:} The simplest approach to compute similarity between two strings which is widely used in information retrieval task is the cosine similarity. Intuitively,  it computes the dot product between the vector representation of the strings. 
\item \textbf{Li's method \cite{li2006sentence}}\footnote{Implementation API available at http://semantic-similarity.azurewebsites.net}: computes the similarity that takes into account the semantic information and word order information. It hinges on using structured lexical database and corpus statistics to compute the semantic similarity.
\item \textbf{DKPro} \cite{bar2013dkpro}\footnote{https://dkpro.github.io/dkpro-similarity/}: Similarity comprises a wide variety of measures ranging from ones based on simple n-grams and common subsequences to high-dimensional vector comparisons and structural, stylistic, and phonetic measures. In our experiment, we use n-gram and Jaccard similarity measures to compute the similarity between two strings.
\item \textbf{Siamese} \cite{neculoiu2016learning}: The method of \cite{neculoiu2016learning} uses character level bi-directional LSTM combined with the Siamese architecture to learn a fixed dimensional representation from variable length sequences using only information about the similarity between pairs of strings. 

\end{itemize}
Comparison with other state-of-the-art STS benchmark models such as the winning model of ECNU team \cite{kutiyanawala2018towards} is not possible as many of them are supervised. This means they need similarity score in [0,5] for training data which is unavailable in the real-world scenario. Note that the SLFR algorithm is an online algorithm whereas Cosine, Li and Siamese  work on pairs of string.  For training SLFR, we split the data into 2/3 for training and 1/3 for testing. For ranking similarity task, we report \emph{precision} (as a performance metric)  of the more similar string with respect to the less similar string. The precision of all of these methods is computed by the fraction of more similar ($s, s_j$) strings returned by them.

\subsection{Empirical Results} In this section, we showcase the comparative evaluation of the SLFR algorithm.  
\subsection{Learning behavior of SLFR}
\begin{figure*}
\subfloat[Invoice]{\includegraphics[width=4.5cm, height=4cm]{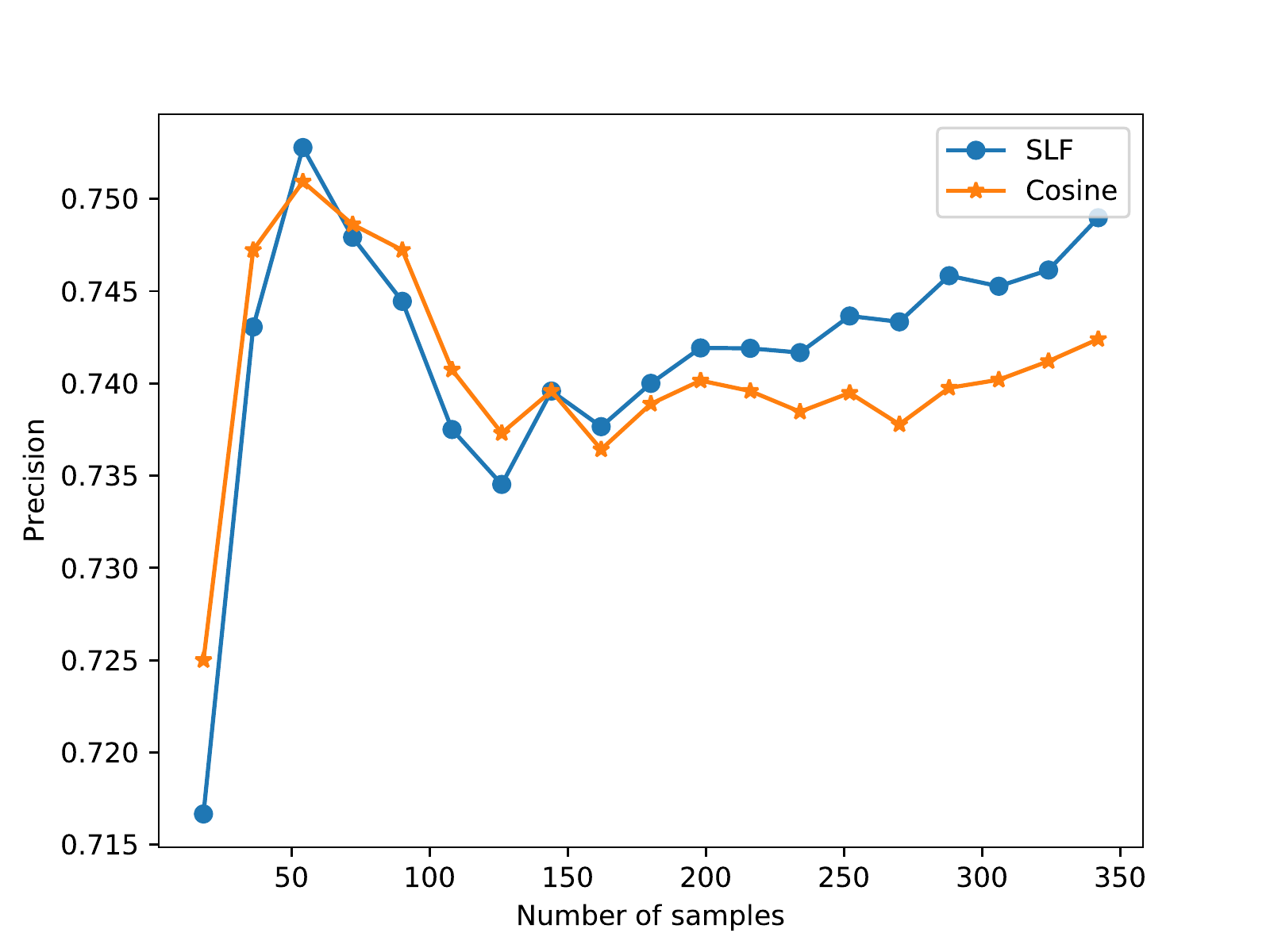}}
\subfloat[Amazon Electronics]{\includegraphics[width=4.5cm, height=4cm]{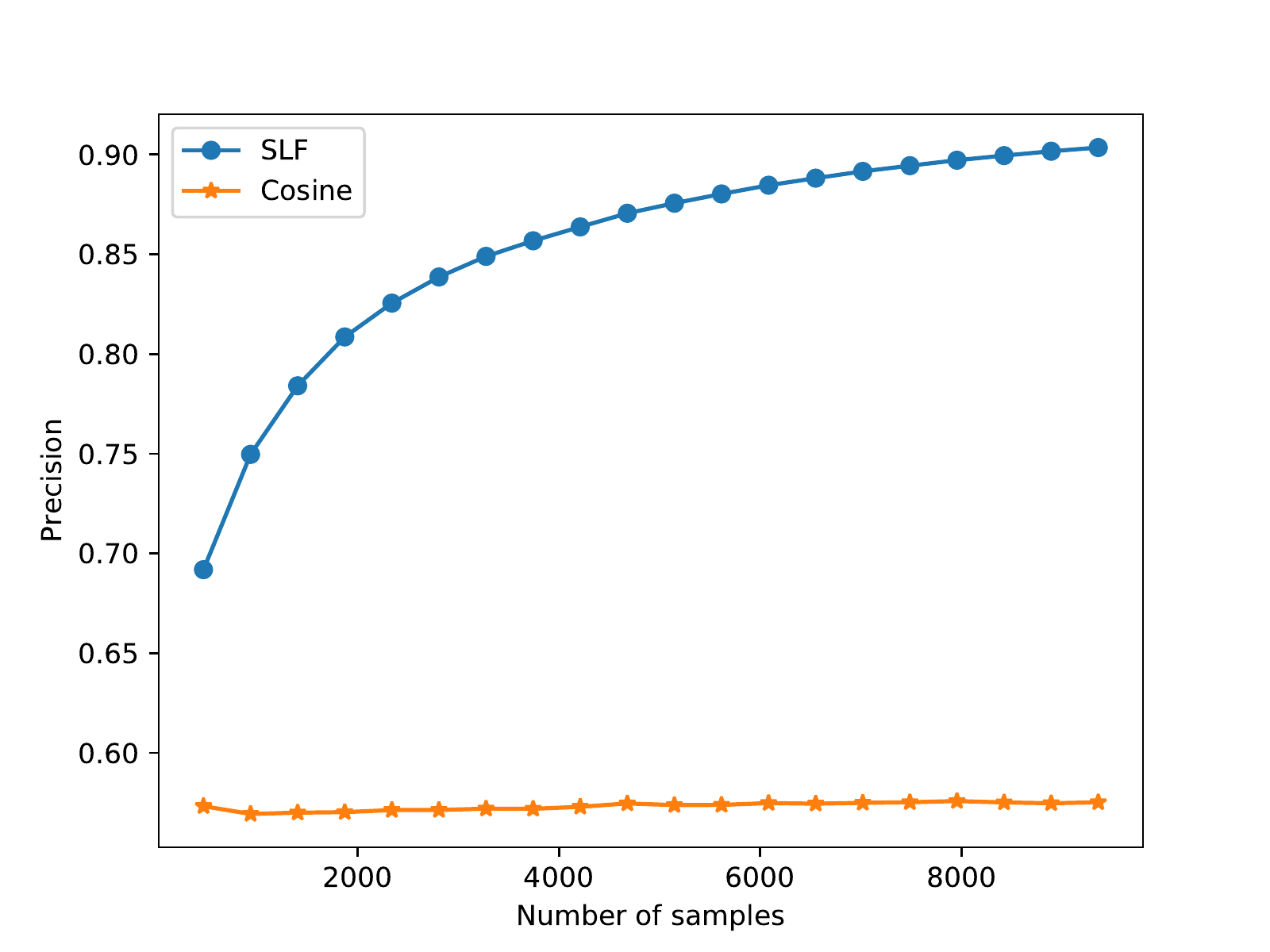}} 
\subfloat[Amazon Automotive]{\includegraphics[width=4.5cm, height=4cm]{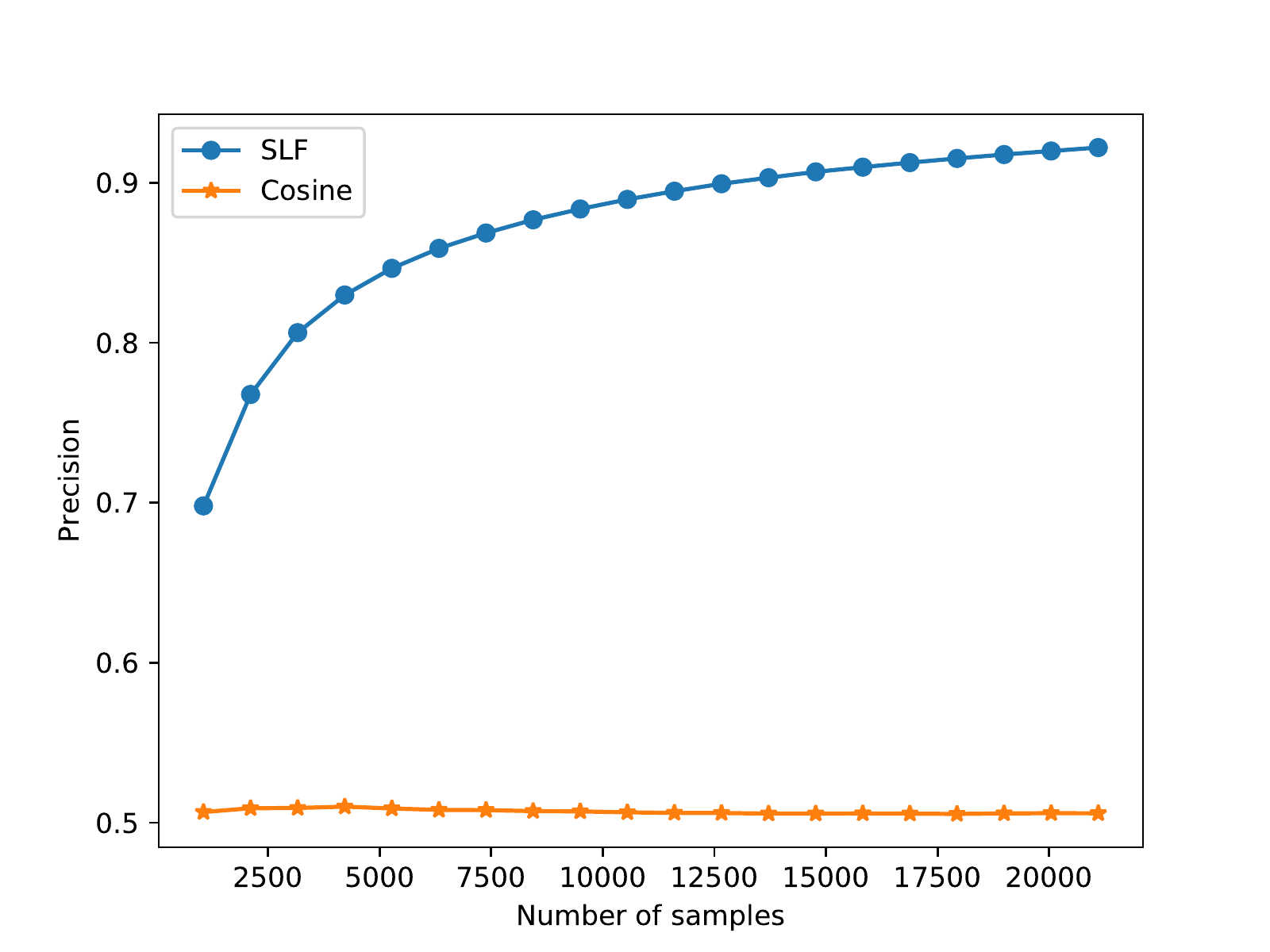}}
\subfloat[Amazon Home]{\includegraphics[width=4.5cm, height=4cm]{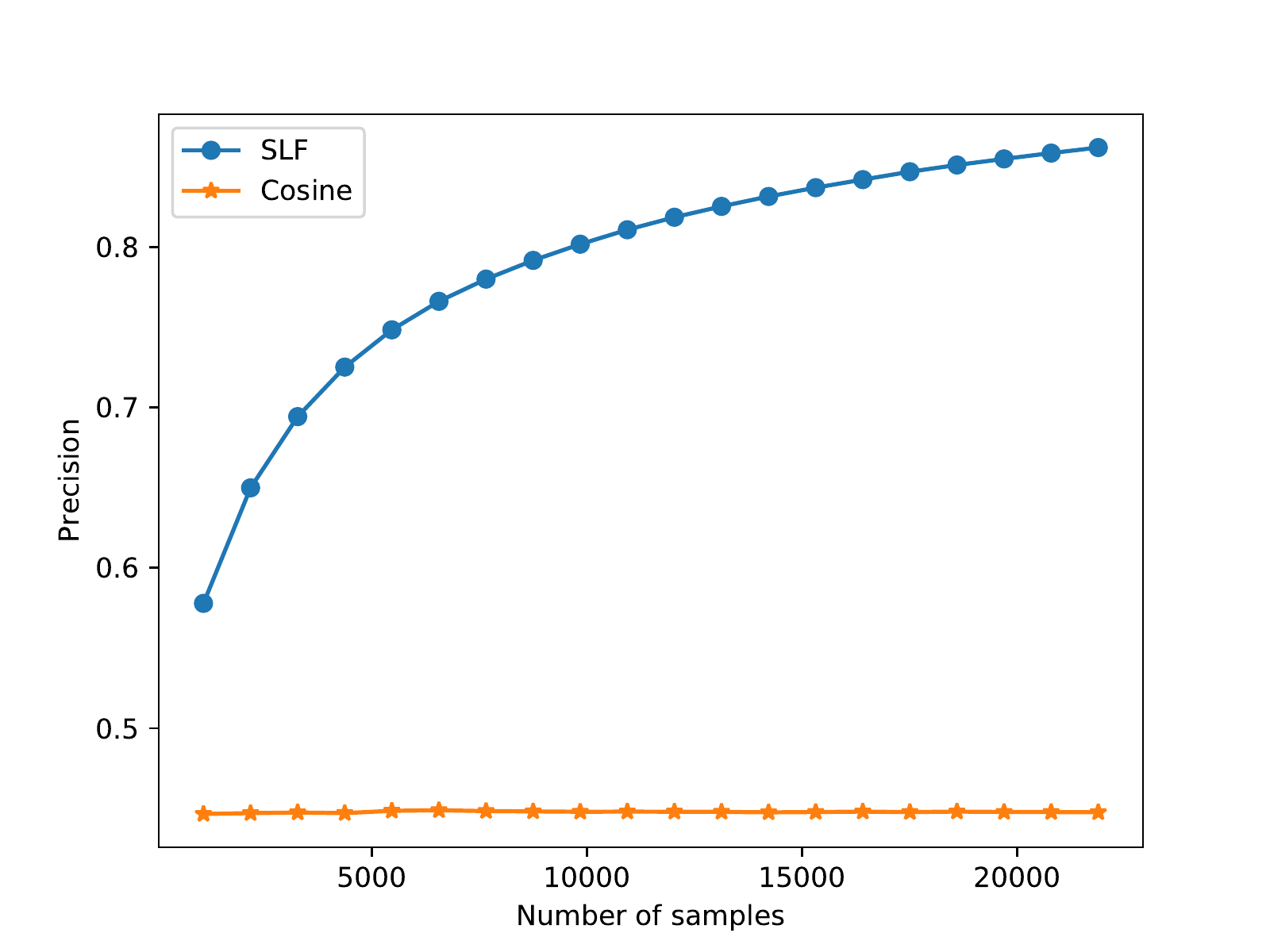}} \\
\subfloat[Flipkart]{\includegraphics[width=4.5cm, height=4cm]{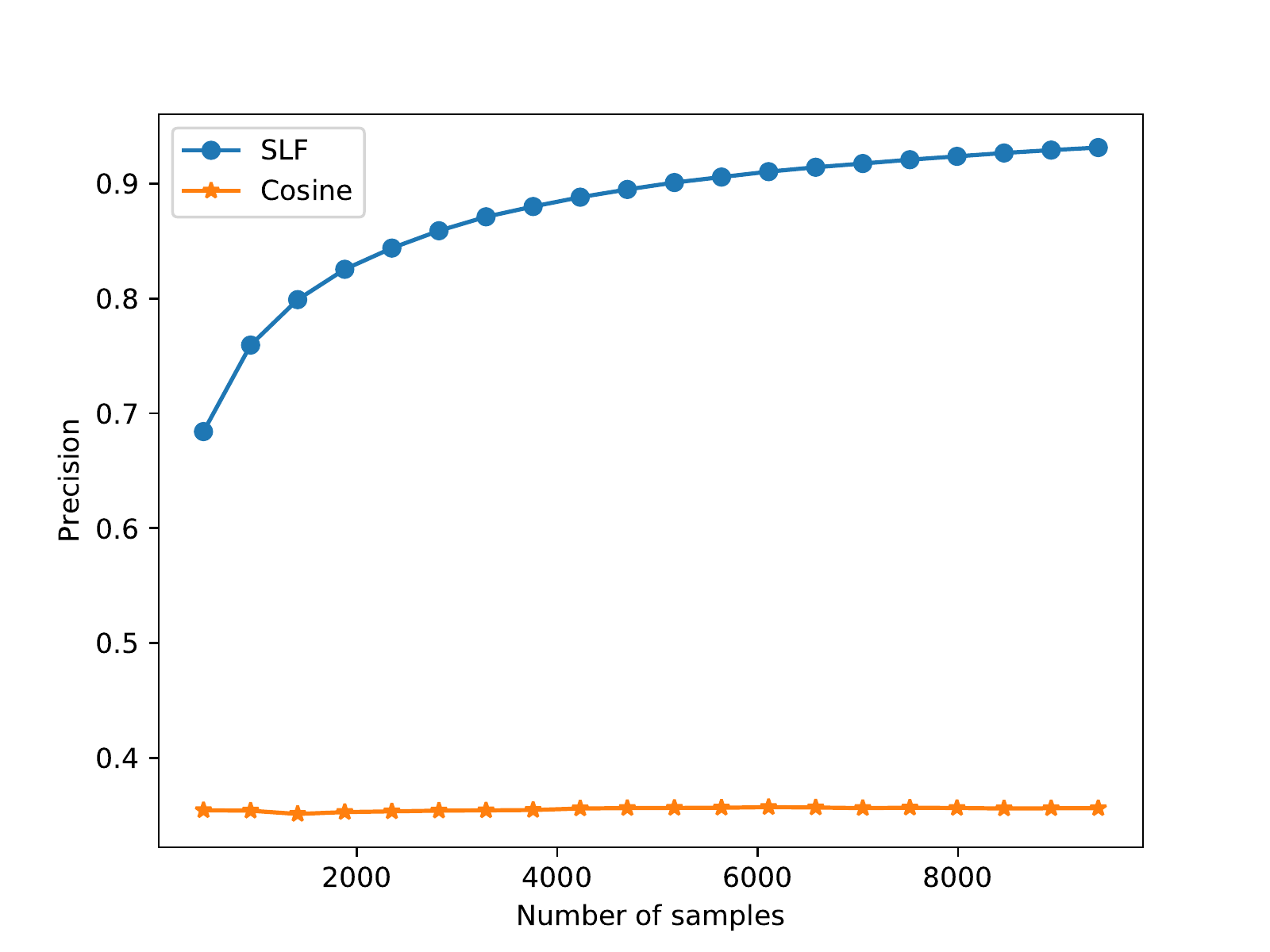}}
\subfloat[SNLI]{\includegraphics[width=4.5cm, height=4cm]{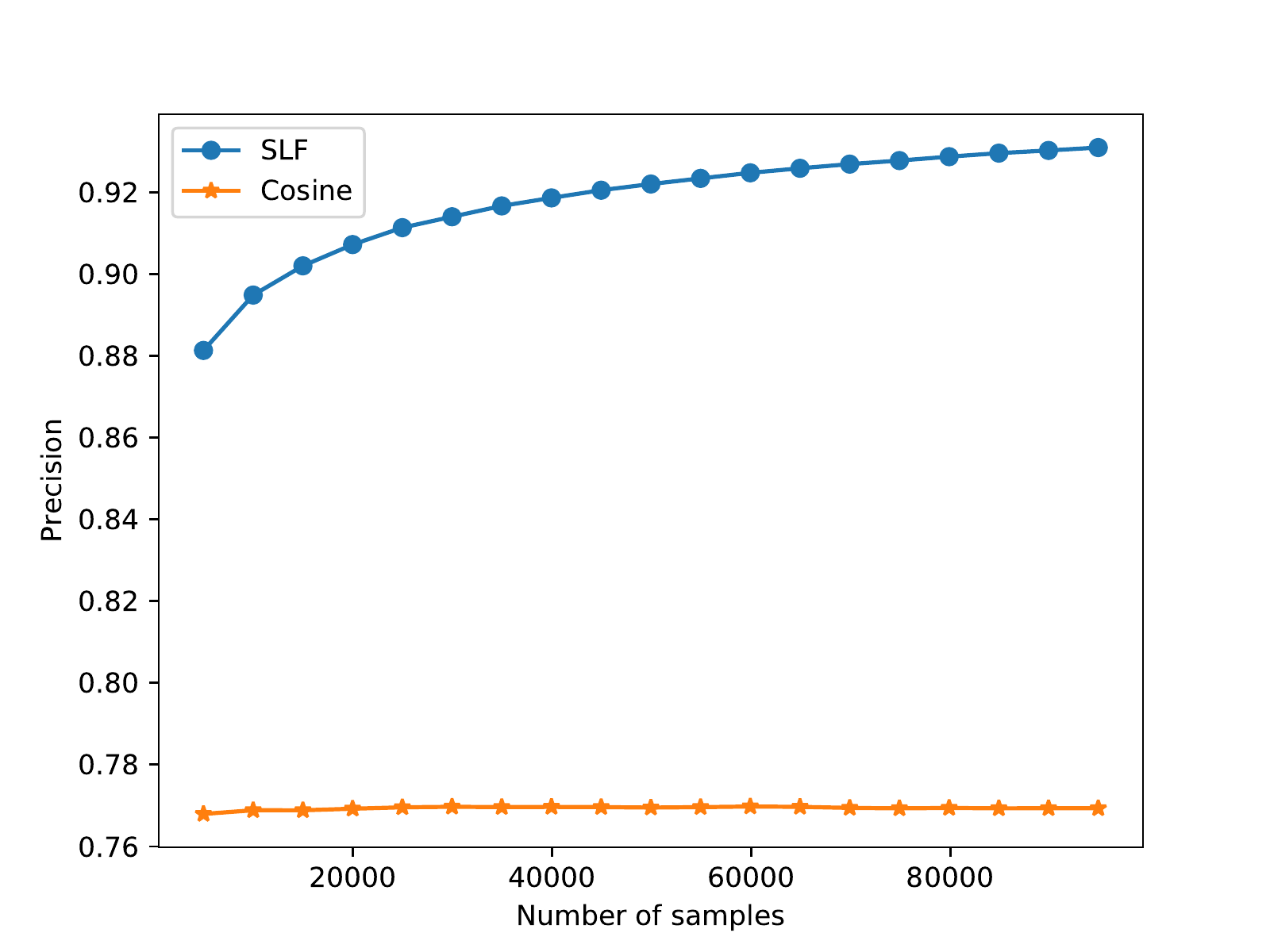}}
\subfloat[SICK]{\includegraphics[width=4.5cm, height=4cm]{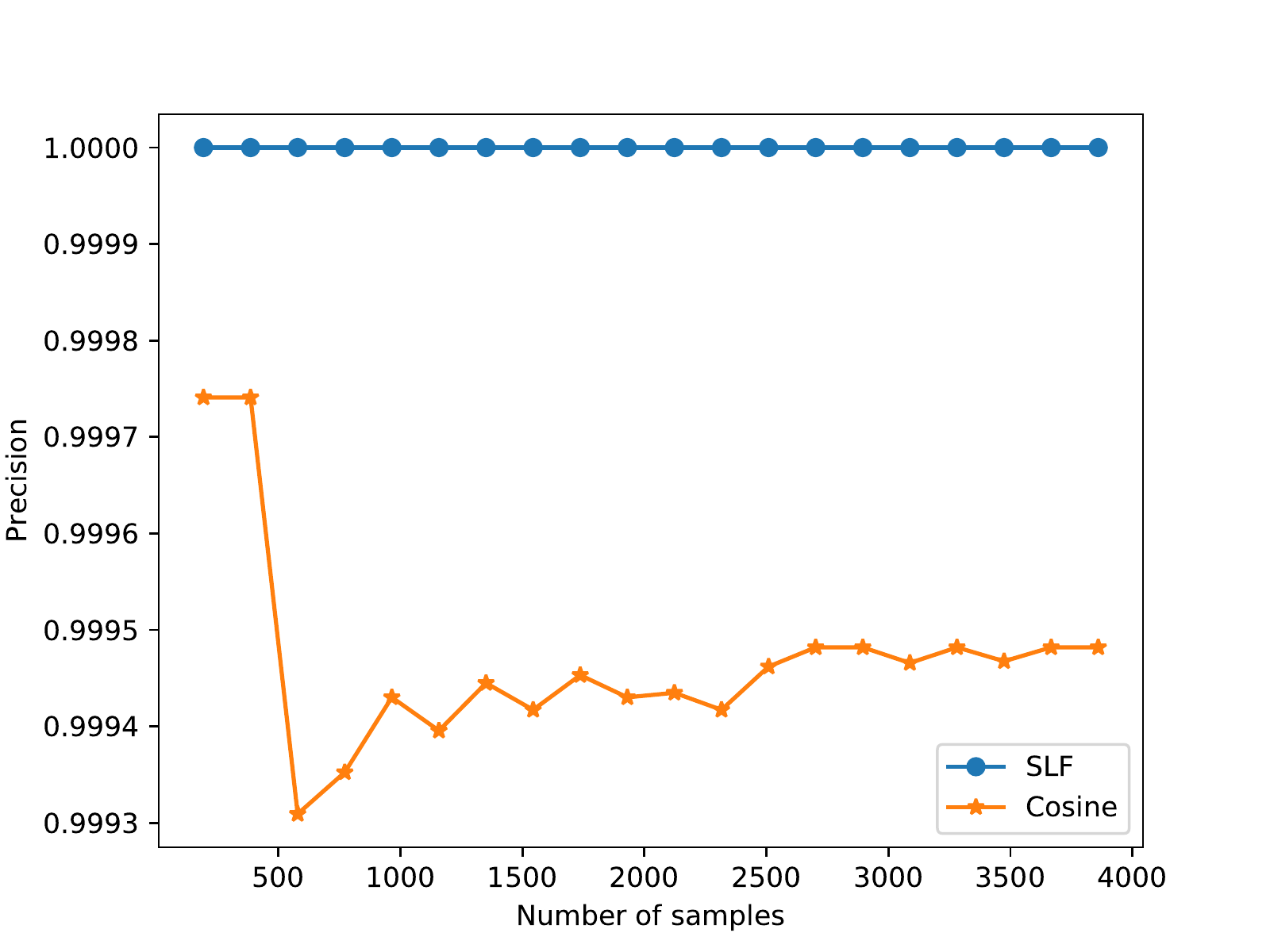}} 
\subfloat[STS]{\includegraphics[width=4.5cm, height=4cm]{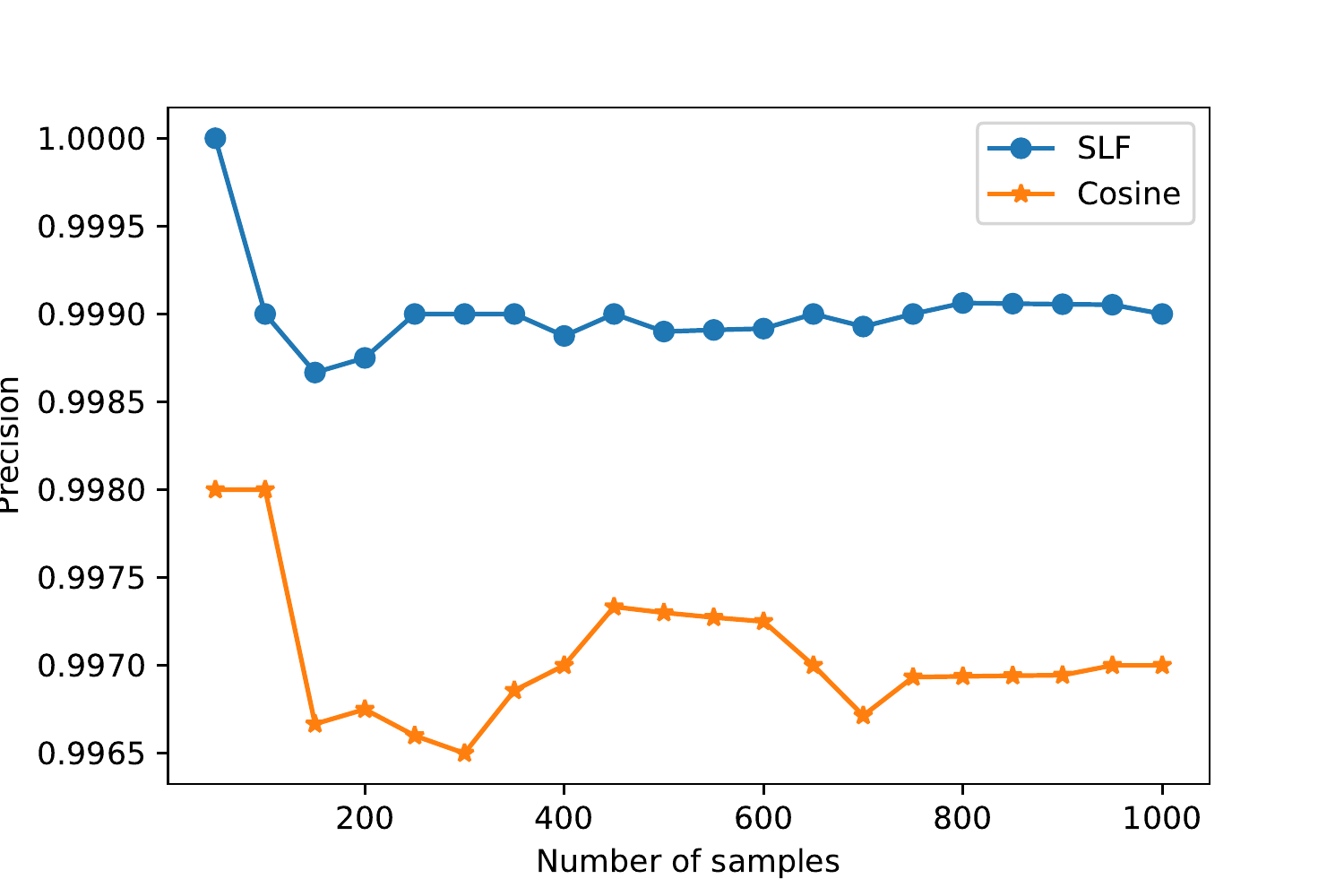}}
\caption{Evaluation of online average of \emph{precision} over 20 random permutation of the training data encoded using tf-idf (a) invoice (b) Amazon electronics (c) Amazon Automotive (d) Amazon Home (e) Flipkart (f) SNLI (g) SICK (h) STS } 
\label{fig:tfidf}
\end{figure*}

\begin{figure*}
\subfloat[Invoice]{\includegraphics[width=4.5cm, height=4cm]{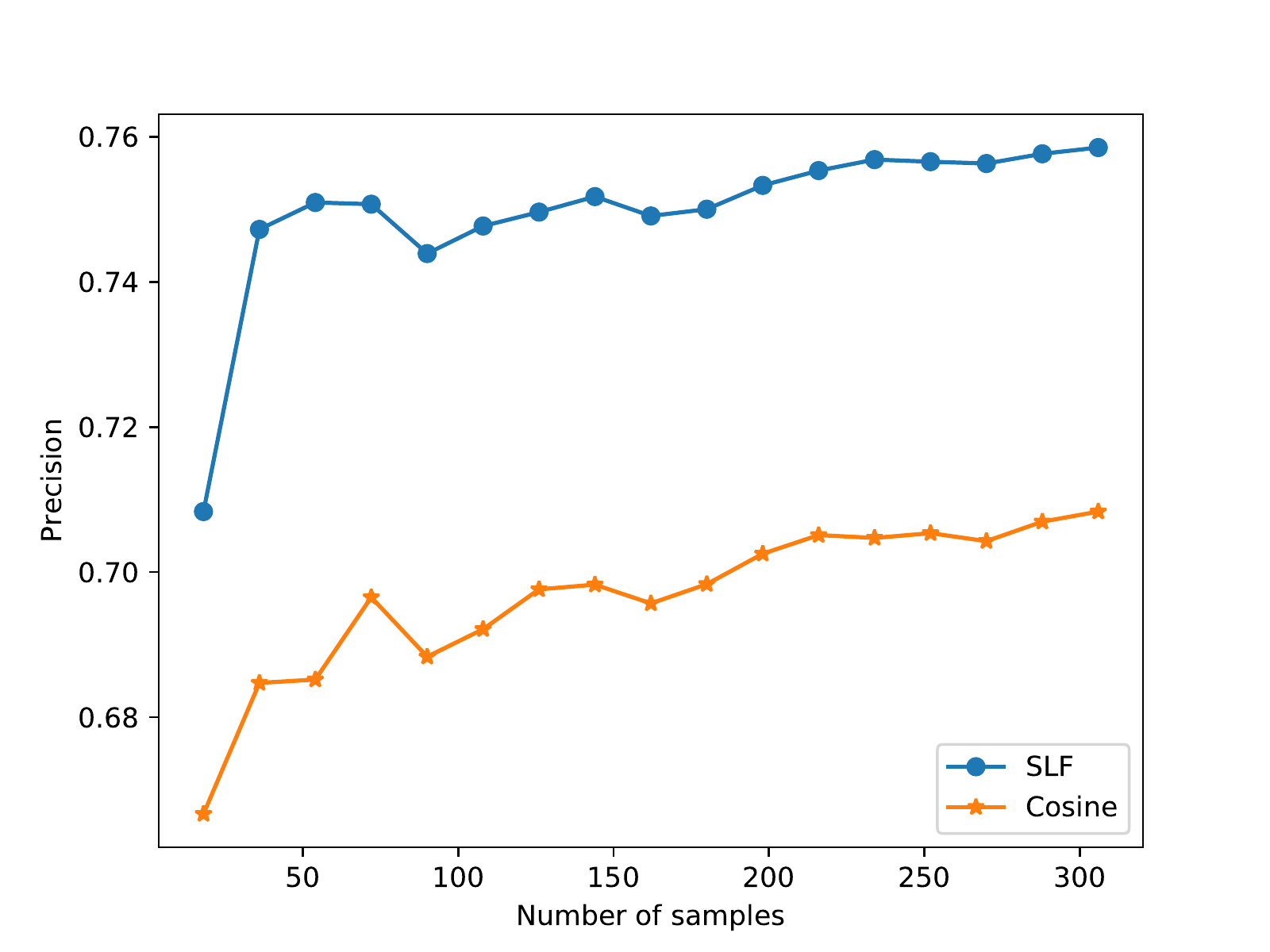}}
\subfloat[Amazon Electronics]{\includegraphics[width=4.5cm, height=4cm]{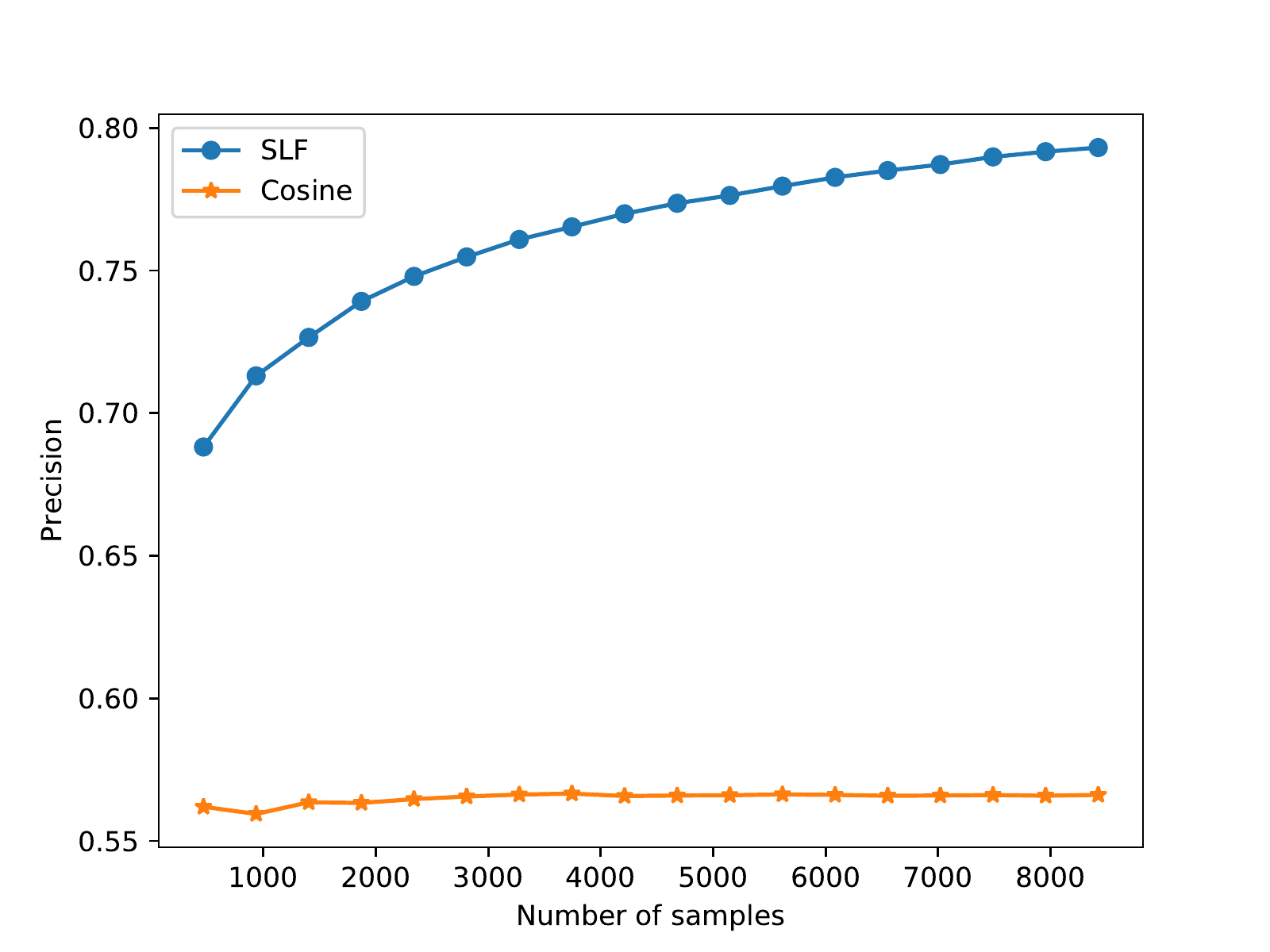}} 
\subfloat[Amazon Automotive]{\includegraphics[width=4.5cm, height=4cm]{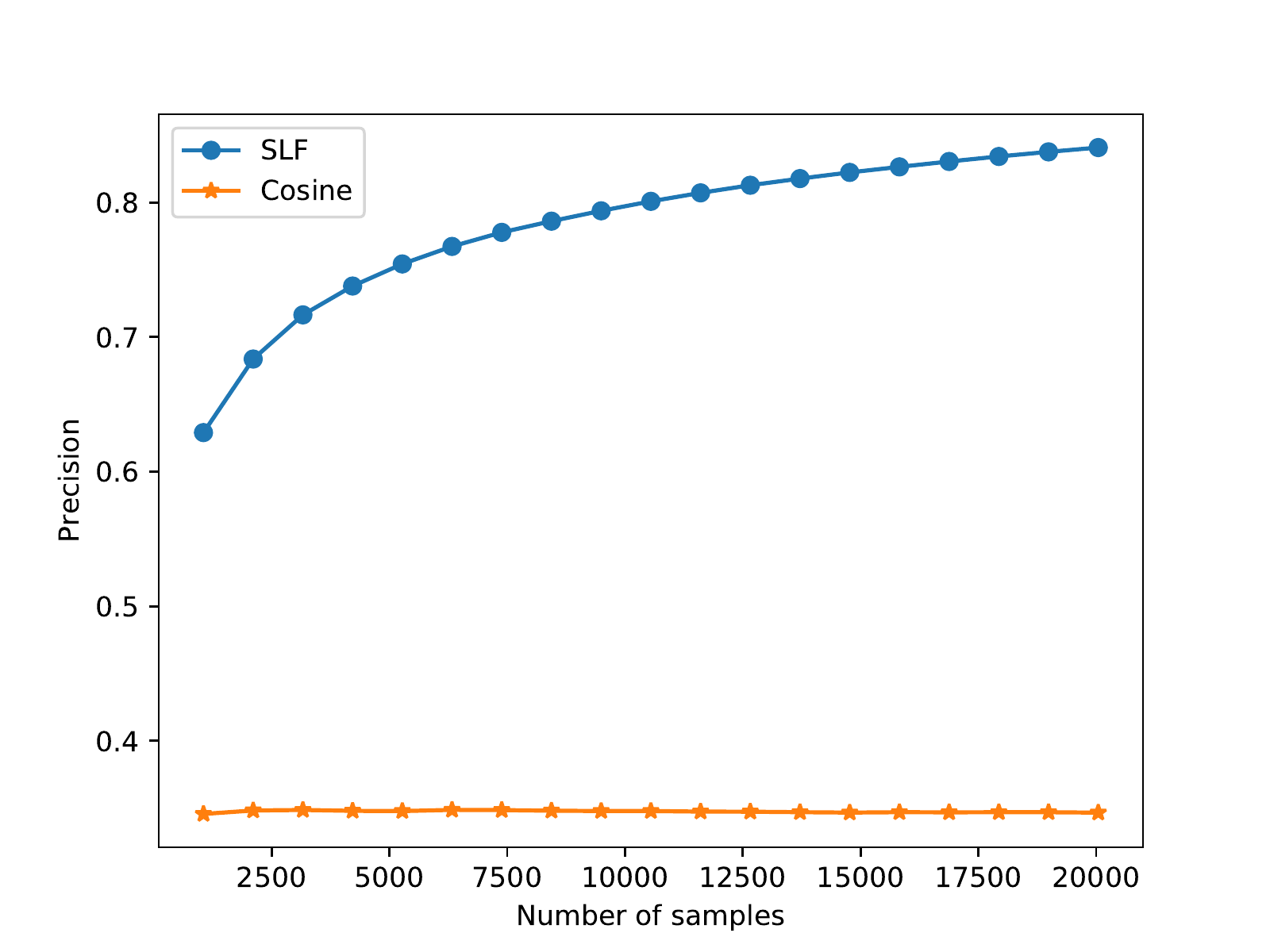}}
\subfloat[Amazon Home]{\includegraphics[width=4.5cm, height=4cm]{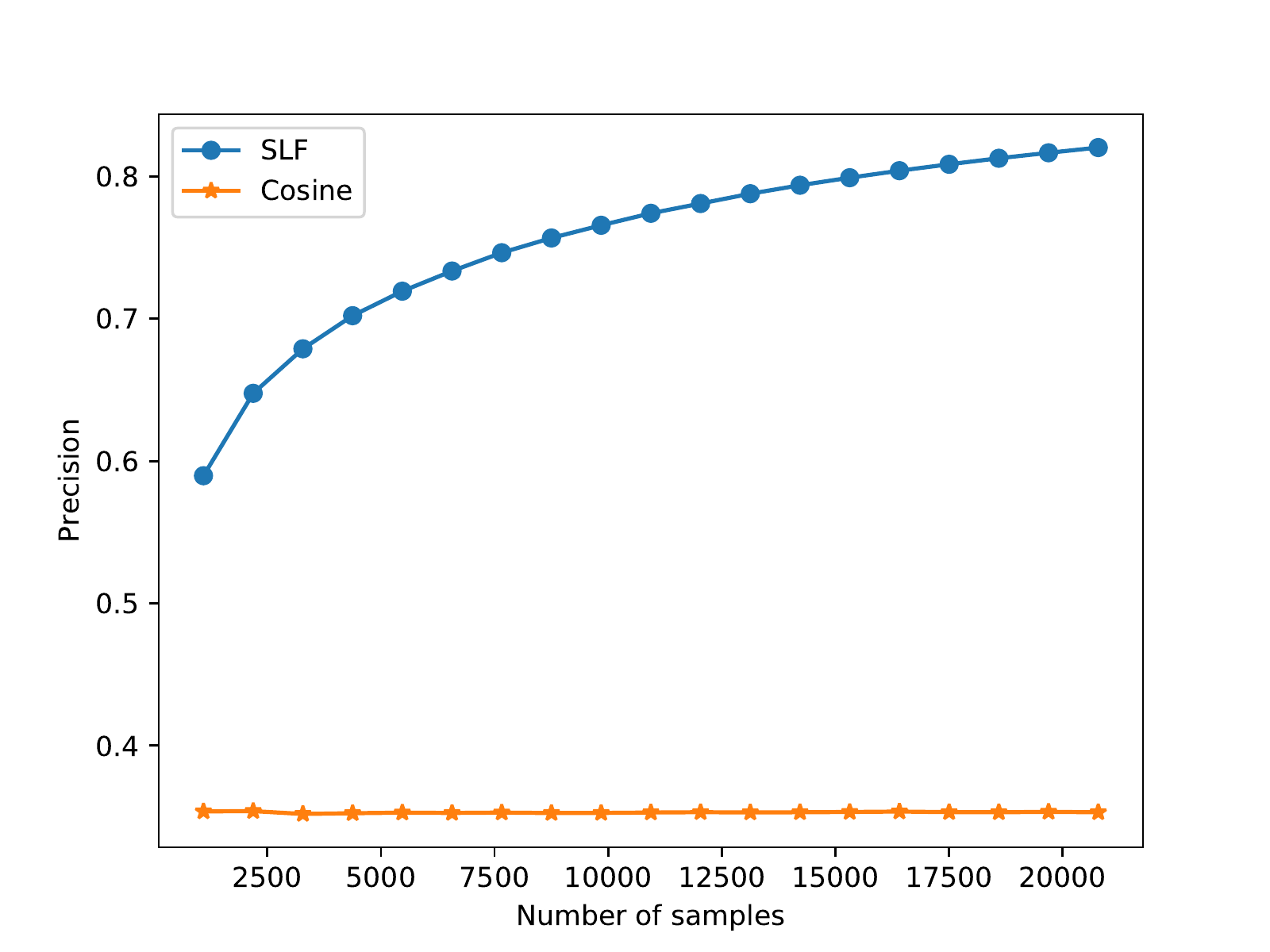}} \\
\subfloat[Flipkart]{\includegraphics[width=4.5cm, height=4cm]{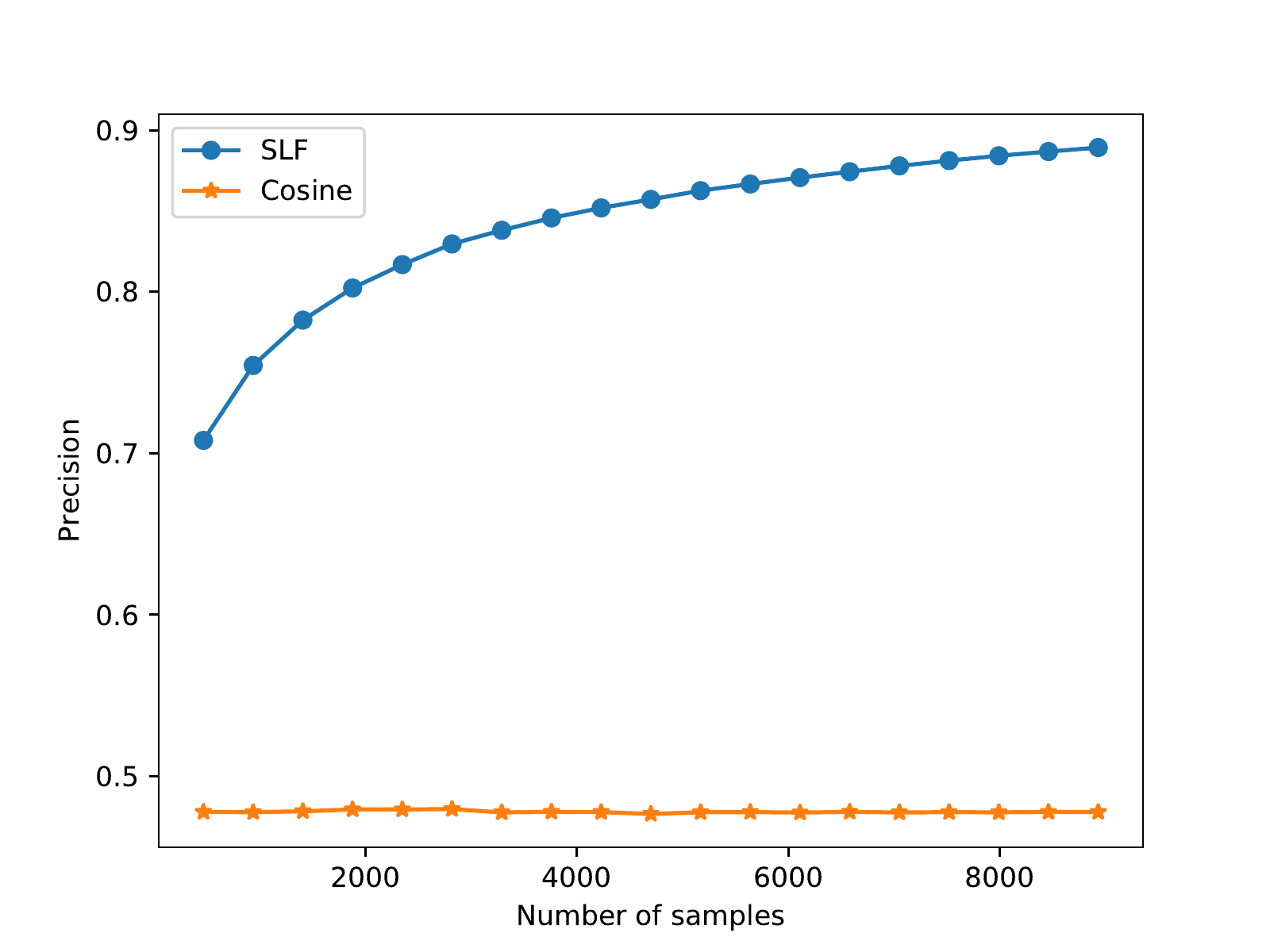}}
\subfloat[SNLI]{\includegraphics[width=4.5cm, height=4cm]{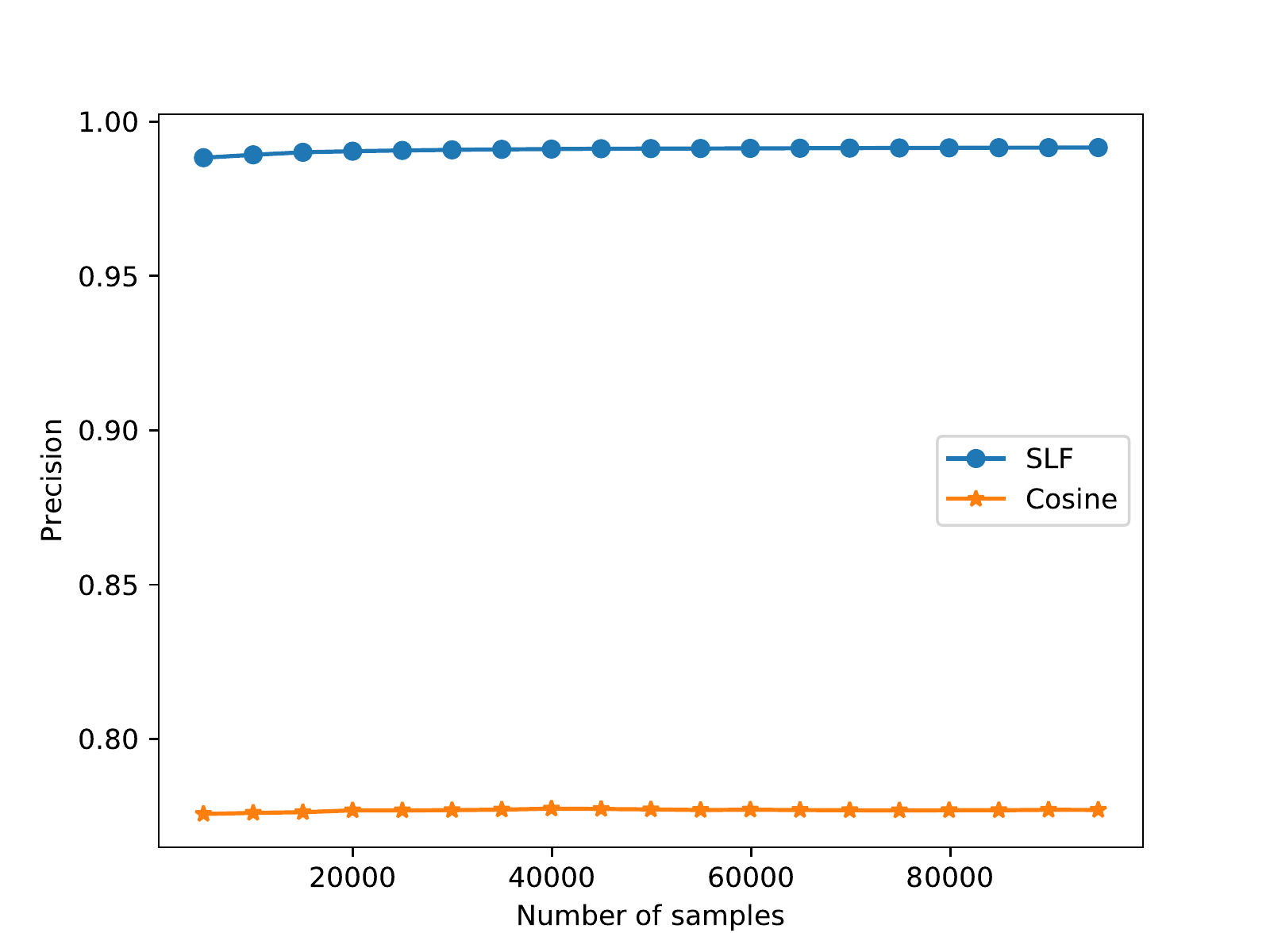}}
\subfloat[SICK]{\includegraphics[width=4.5cm, height=4cm]{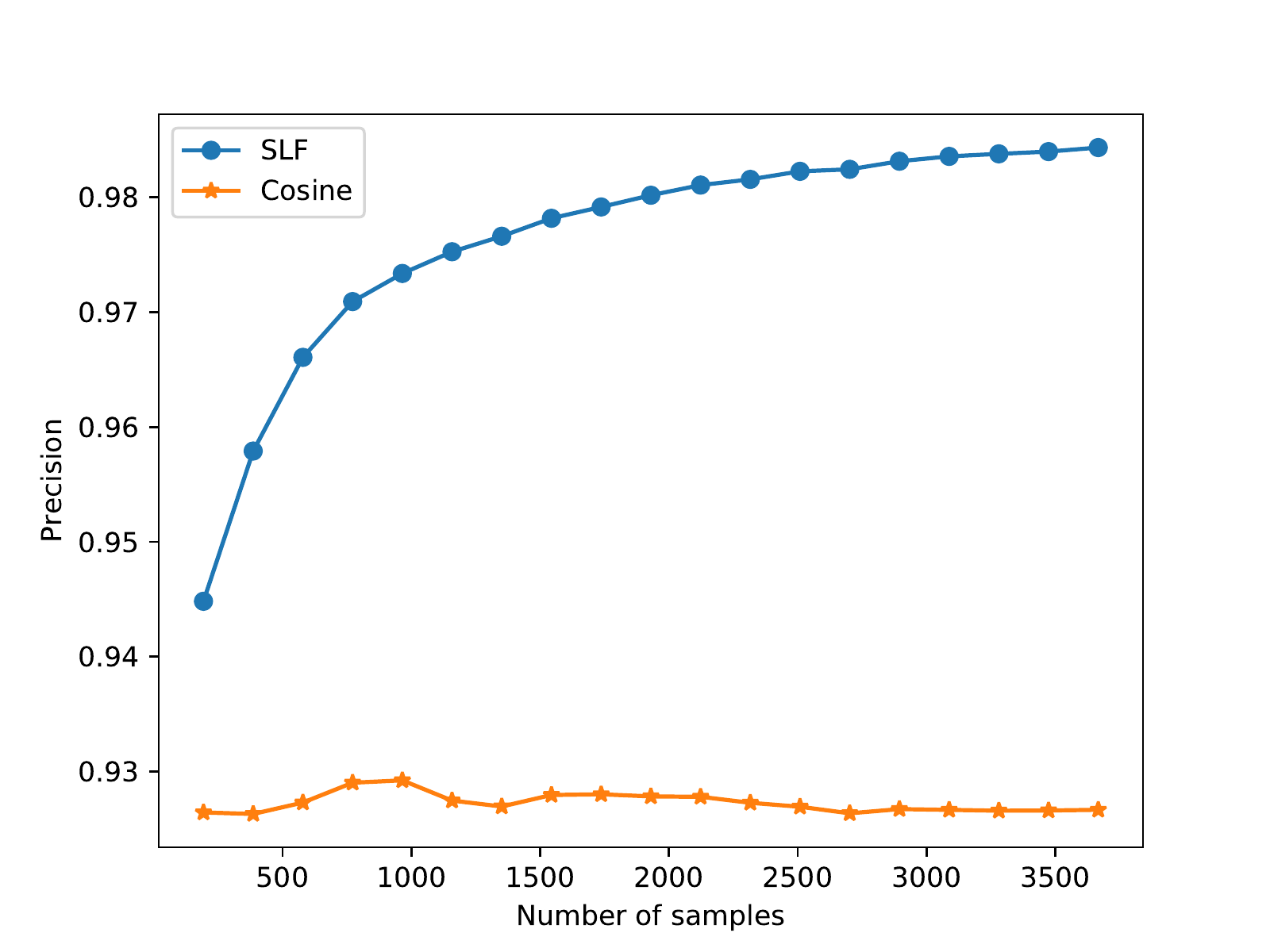}} 
\subfloat[STS]{\includegraphics[width=4.5cm, height=4cm]{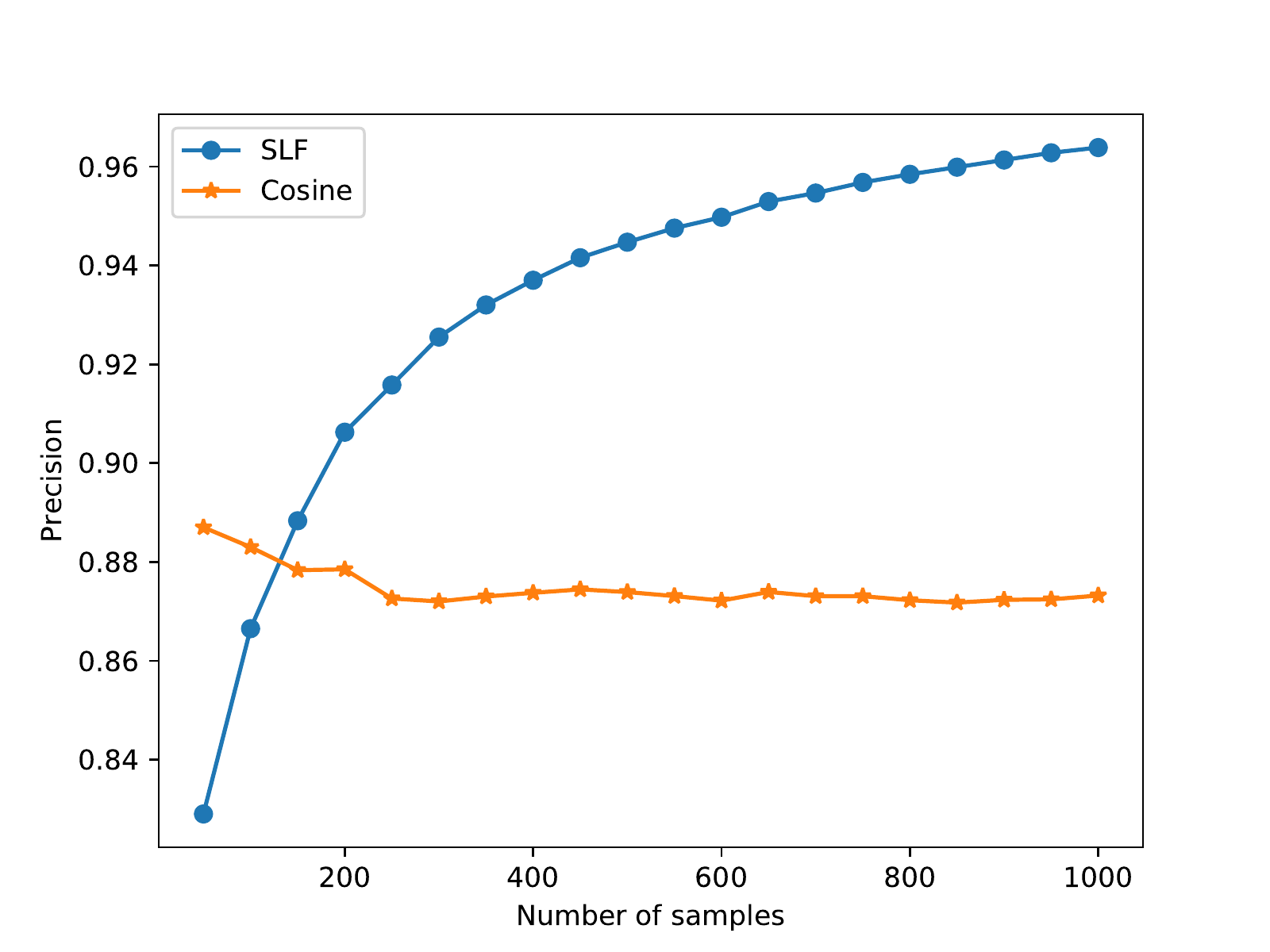}}
\caption{Evaluation of online average of \emph{precision} over 20 random permutation of the training data encoded using Infersent. (a) invoice (b) Amazon electronics (c) Amazon Automotive (d) Amazon Home (e) Flipkart (f) SNLI (g) SICK (h) STS } 
\label{fig:infer}
\end{figure*}

In this section, we present the learning behavior of the SLFR with respect to the number of samples seen. We also discuss the online average \footnote{online average refers to the average result collected at different points of sample sizes as it arrives.} of precision over 20 random permutations of the training data using tf-idf, Infersent and USE encoding of the strings. The results are shown in Fig. \ref{fig:tfidf} and  \ref{fig:infer}  (USE encoding results not shown due to lack of space) respectively. We can draw several conclusions from it. Firstly, from Fig. \ref{fig:tfidf}, we can see that SLFR approach to learn the similarity between invoice and PO string is consistently achieving higher precision compared to vanilla cosine similarity on all the datasets. Secondly, since there is no learning mechanism involved in the cosine, its performance is fluctuating over the number of training samples which is obvious. Thirdly, there is an initial performance drop for the SLFR on invoice and STS datasets. It could be possibly due to the more complex nature of the strings in these datasets which brings some noise during the preprocessing. Surprisingly, learning behavior of the SLFR on SICK data remains stagnant. 

Empirical evaluation of precision using Infersent encoding is shown in Fig. \ref{fig:infer}. Interestingly, in Fig. \ref{fig:infer}, we can clearly see that SLFR is outperforming compared to cosine on \emph{all} the datasets when we encode strings using Infersent. Besides, learning behavior of SLFR is smoother using Infersent encoding compared to tf-idf encoding scheme (visible in invoice and STS datasets). However, the precision achieved using tf-idf encoding scheme is better than Infersent in most of the cases which, for example, can be seen on \emph{Amazon Electronics} dataset. Evaluation of precision using USE encoding scheme is not shown due to lack of space but it follows similar trend as that of Infersent.  Further, if we compare precision achieved using three encoding schemes on a particular dataset, for example, \emph{Amazon Electronics}, USE scheme performs worst while the tf-idf scheme performs the best. This could be attributed partially due to OOV word removal during preprocessing. Training USE/Infersent on the domain-specific data can surely lift the precision. However, such a study is not conducted due to non-availability of the large-scale \emph{ranked data} from the e-commerce domain required for training USE/Infersent models and left as a future study.

\subsection{Generalization Performance of SLFR}
In the previous section, we studied the behavior of the SLFR on the training data using the three encoding schemes. In this section, we study the behavior of the proposed algorithm on the test data to understand the generalization performance using the three encoding scheme. The results are shown in Table \ref{encoding}. From the results in the table, we can clearly see that the tf-idf encoding scheme is competitive and achieves better average precision on most of the datasets. As such, we use tf-idf character n-gram encoded data to show the comparative evaluation with respect to the baselines.

\begin{table}[]
\caption{Evaluation on the test data (average precision  over 20 runs ) using three encoding schemes}
\label{encoding}
\centering
\begin{tabular}{|l|l|l|l|}
\hline
 \backslashbox[]{Data}{Encoding}   & Tf-idf & Infersent & USE   \\ \hline
Invoice                                            &\textbf{0.77}  & 0.62     & 0.62 \\ \hline
Amazon Electronics                                 & \textbf{0.80}  & 0.65     & 0.75 \\ \hline
Amazon Automotive                                  & \textbf{0.90}  & 0.82     & 0.50 \\ \hline
Amazon Home                                        & \textbf{0.86}  & 0.82     & 0.77 \\ \hline
Flipkart                                           & 0.86  & \textbf{0.87}     & 0.84 \\ \hline
SNLI                                               & 0.94  &  0.94         &\textbf{ 0.99} \\ \hline
SICK                                               & \textbf{0.99}  & 0.95     & 0.99 \\ \hline
STS                                                &\textbf{0.99}  & 0.95     & 0.98 \\ \hline
\end{tabular}
\end{table}

\subsection{Comparative Performance Evaluation of SLFR}
In this section, comparative evaluation of various algorithms is presented as shown in the Table \ref{tab:recall}. We compare the average precision over 20 random permutation of the test data. From the result in the table, we can see that SLFR achieves comparable or superior performance on most of the datasets. Interestingly, Siamese architecture shows poor performance. The reason could be attributed to the fact that it is not able to catch small variations in the input strings. The other reason could be that the score assigned by it to the similar pair is same as the dissimilar pair and our score calculation mechanism  says that precision will go up only when there is a strict inequality, i.e., $score(s,s_j) \geq score(s,s_i)$ only then it will add to the precision. We suspect both of the reasons behind the low performance of Siamese architecture.

\begin{table}
\tiny
\caption{Comparative evaluation of average precision in \% on test data }
\label{tab:recall}
\begin{tabular}{|p{2cm}|l|l|l|l|p{1cm}|}
\hline
\backslashbox[]{Data}{Algo}  & SLFR  & Cosine & Li    & DKPRo & Siamese \\ \hline
Invoice                                            &\textbf{0.77} & 0.73  & 0.65 & 0.52 & 0.13   \\ \hline
Amazon Elec.                                & \textbf{0.80} & 0.58  & 0.40 & 0.41.34 & 0.09    \\ \hline
Amazon Auto.                                 & \textbf{0.90} & 0.51  & 0.39 & 0.40 & 0.09    \\ \hline
Amazon Home                                        & \textbf{0.86} & 0.45  & 0.40 & 0.35 & 0.10   \\ \hline
Flipkart                                    & \textbf{0.86} & 0.28  & 0.24 & 0.27 & 0.05    \\ \hline
SNLI                                               & \textbf{0.94} & 0.76  & 0.75 & 0.20 & 0.23   \\ \hline
SICK                                               & \text{0.99} & \textbf{0.99}  & \textbf{0.99} & 0.79 & 0.09   \\ \hline
STS                                                & \textbf{0.99} & \textbf{0.99}  & \textbf{0.99}& 0.79 & 0.08    \\ \hline
\end{tabular}
\end{table}

 \begin{table}[b]
\caption{An example of line items matching}
\label{validity}
\centering
\begin{tabular}{|l|l|}
\hline
Invoice                           & PO                                          
\\ \hline \hline
Edible oil 5 lt  & \begin{tabular}[c]{@{}l@{}}1. Coconut oil 2 lt\\ 2. Sunflower oil 2 lt \\ 3. Musterd oil 1 lt \end{tabular} \\ \hline
Diesel oil 5 lt  & \begin{tabular}[c]{@{}l@{}} Diesel oil 5 lt\end{tabular}                   \\ \hline
\end{tabular}
\end{table}

\section{Threats to Validity} \label{threats}
The approach proposed in the previous section to match invoice and PO line items will fail in the situation when there are multiple line items in PO which match a single item in the invoice. In other words, there is a hierarchical relation between the products in the invoice and PO. Secondly, it is cumbersome for agents to manually club the matching items from the PO. 

 An illustrative example is shown in Table \ref{validity}. As we can see that \quotes{Edible oil 5 lt} in the invoice column should match the oils listed in the PO column. However, the approach presented in the previous section fails to cater to such lines items even if the agent provides relative feedback. Another solution to handle these hierarchical relationships may be to use hypernyms/hyponyms relation available in the lexical databases such as Wordnet. However, using Wordnet to resolve this issue may not be sufficient.  Hence, we propose a subroutine in the next section to handle such cases.

\section{Using Product Taxonomy and Catalogue for Matching Line Items}
\label{prodtaxo}
The proposed approach uses the product taxonomy to resolve the issue of the hierarchical relationship among the line items. In addition, it is also shown how can we leverage product catalogs if available for matching line items.

\begin{figure}
\subfloat[]{
    \includegraphics[width=3cm, height=3cm]{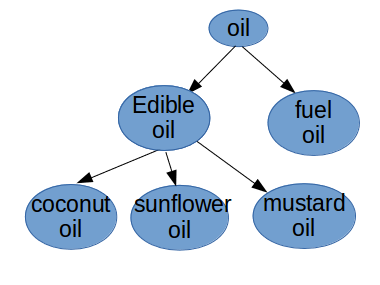}}
\subfloat[]{
\includegraphics[width=5cm, height=4cm]{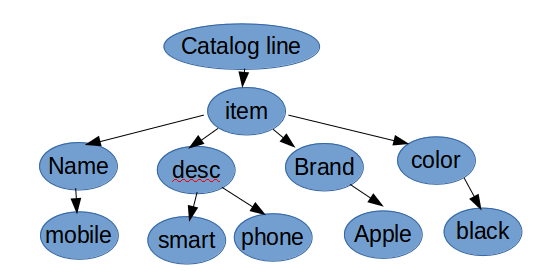}}   
    \caption{(a) product taxonomy (b) product catalogue}
    \label{fig:taxonomy}
\end{figure}

A typical product taxonomy is shown in Fig. \ref{fig:taxonomy}(a). A product taxonomy clearly depicts the hierarchical relationship among the products. For example, In Fig. \ref{fig:taxonomy}(a), it shows that \emph{coconut oil} is an instance of the \emph{edible oil} which in turn is an instance of the more general concept \emph{oil}. Such a categorization of products is generally available with the e-commerce vendors. In other cases where it is not available, an \emph{ontology}\footnote{a set of concepts and categories in a subject area or domain that shows their properties and the relations between them} can be learned given sufficient data from the concerned domain. It is assumed that such an ontology is available and taxonomy is a part of the more complex ontology.
\begin{algorithm2e}[tb]
	\SetAlgoLined\DontPrintSemicolon
	{\small
		\KwIn{invoice and PO line items}{
		\KwOut{Similarity score}
	\Begin{	
   \label{algo2}
      \begin{enumerate}
          \item  Extract product names from PO and Invoice using catalogues(if available) or via  (NER).\;
          \item
Check if the item in the invoice is a generalization of \;an item(s) in the 
PO using product taxonomy.\;
        \item If  step 2 returns True,  combine all the item(s) in PO\; and make it as one line item.
        \item  Extract product attributes present in the catalogue. 
        \item  Do the fuzzy matching of the entities and attributes\; in the invoice  line description. Suppose, out of $n$ entities/attributes, $k_1$ entities/attributes cross a \;threshold in the invoice line item.
        \item Run NER model to extract product attributes from\; PO line item. Let $k_2$ be the set of such entities.
        \item Check for hierarchal relationships of entities present\; in the sets $k_1$ and $k_2$ using taxonomy.  Let $k$ be the set \;of tokens with existing relationship. 
\item Next, merge the tokens which are common to the \;invoice and PO line items to $k$, call it $k^*$.
\item Build the set $k_3$ of the remaining tokens. 
\item Compute the Jacccard similarity by 

\begin{equation*}
Jsim = \frac{|k^*|}{|k_1 \cup k_2 \cup k_3-k|}
\end{equation*}
\item return \emph{Jsim}
      \end{enumerate}
   }
   }
   }
	\caption{Matching Line Items using Taxonomy and Catalog}\label{algo2}
\end{algorithm2e}
We also use the product catalog corresponding to the invoice line item as shown in fig. \ref{fig:taxonomy}(b). Note that one may think that the product catalog contains all the information about the product. Though it is true that the invoice line item \emph{may} contain more information than just available in the catalog. For example, \emph{5 boxes of Blak Appel Mobile 6s 32 Gb memry }. As we can see that \emph{5 boxes} will not be present in the catalog corresponding to the \emph{apple mobile}. This is the reason we can not solely rely on matching catalogs corresponding to the invoice and PO line items.

Now, we have product taxonomy and catalog for line item matching. Pseudocode for line item matching using product taxonomy and catalog is given in the Algorithm \ref{algo2}. We avoid further description of the Algorithm \ref{algo2} due to lack of space. 

\section{Conclusion and Future Work}\label{future}
Invoice line item description matching is a resource-intensive task in procure to pay business process.  We proposed two approaches using the domain expert's feedback on description matches. We showed that using domain knowledge to learn similarity ranking or classification models outperforms existing approaches.  By employing our solution, invoice processing cost and time can be reduced significantly. In future work, we would like to reduce the number of agent's feedback.  We are also exploring the possibility of extending the work towards matching vendor names and address across invoices and PO.
\section*{Acknowledgement}
Project no. ED\_18-1-2019-0030 (Application-specific highly reliable IT solutions) has been implemented with the support provided from the National Research, Development and Innovation Fund of Hungary, financed under the Thematic Excellence Programme funding scheme.
\bibliographystyle{aaai}
\bibliography{main}
\end{document}